\documentclass[letterpaper, 10 pt, conference]{ieeeconf}  

\IEEEoverridecommandlockouts                              

\usepackage{subcaption}
\usepackage{graphicx}
\usepackage{comment}
\usepackage{color}
\usepackage{amsmath}
\usepackage{amssymb}
\usepackage{multirow}
\usepackage{tabularx}
\usepackage{threeparttable}
\usepackage{booktabs}
\usepackage{makecell}
\usepackage{adjustbox}
\usepackage{pifont}
\usepackage[table]{xcolor}

\title{\LARGE \bf
ProCal: Inference-Time Proposal Calibration for Open-Vocabulary Object Detection
}

\author{Jae-Ryung Hong$^{1}$, Ho-Joong Kim$^{1}$, and Seong-Whan Lee$^{1}$%
\thanks{*This work was supported by the Institute of Information \& Communications Technology Planning \& Evaluation (IITP) grant funded by the Korea government (MSIT) (No. RS-2019-II190079, Artificial Intelligence Graduate School Program, Korea University), and by the Information Technology Research Center (ITRC) support program (No. IITP-2026-RS-2024-00436857).}
\thanks{
$^{1}$J.-R. Hong, H.-J. Kim, and S.-W. Lee are with the Department of Artificial Intelligence, Korea University, Anam-dong, Seongbuk-ku, Seoul 02841, Korea.
\texttt{\small \{h\_jr, hojoong\_kim, sw.lee\}@korea.ac.kr}
}}

\begin{document}

\maketitle
\thispagestyle{empty}
\pagestyle{empty}

\begin{abstract}

Open-vocabulary object detection aims to localize and classify objects beyond the fixed set of categories seen during training. Recent open-vocabulary object detection methods improve localization and classification for unseen categories by leveraging a frozen VLM as a detector backbone. However, VLM classification score lacks recognizing position and scale of the object in an image. We observe that pretrained VLMs enable to classify foreground and background regions. According to this observation, we propose a simple inference-time Proposal Calibration (ProCal) that improves localization quality of the classification score. ProCal computes a proposal prior by combining two scores: localization-aware foreground score and background-aware suppression score. Localization-aware foreground score captures whether a proposal contains an object area. Background-aware suppression score measures the extent to which the proposal resembles background. We analyze that ProCal suppresses false novel activation on background proposals and consistently ranks true novel proposals above background and partial novel proposals. Applied to CLIPSelf ViT-L/14, ProCal improves $\text{AP}_r$ +2.5 on OV-LVIS. The analyses show that proposal-level localization-aware reranking effects to mitigate ranking miscalibration for novel objects.

\end{abstract}


\section{INTRODUCTION}

Category-level generalization has been studied across zero-shot and transfer-learning recognition settings~\cite{ref1, ref2, ref3}.
Open-vocabulary object detection (OVD) instantiates this issue in object detection, where detectors trained with base-category annotations are required to localize and classify categories that are not annotated during detector training. 
By leveraging text embeddings from pretrained vision-language models (VLMs) such as CLIP~\cite{clip}, OVD enables the localization and classification of unseen objects.
Through pretrained image-text alignment, the detector associates object regions with semantic class descriptions.
Transferring this image-level alignment of pretrained VLMs to region-level recognition for object detection remains challenging.

Previous methods~\cite{ViLD,Detic,RO-ViT,DST-Det} adapt vision-language models to detection through additional pretraining, transfer learning, or fine-tuning to improve region-text alignment.
F-VLM~\cite{F-VLM} incorporates the image encoder of a pretrained VLM into the detector backbone, motivated by the observation that pretrained VLM features capture object boundaries.
Building on this idea, recent open-vocabulary detectors~\cite{CLIPSelf,OV-DQUO} follow this approach that freezes a pretrained VLM as the image backbone.
These F-VLM-style methods~\cite{CLIPSelf,OV-DQUO,CORA} compute VLM scores using the cosine similarity between text-prompt and region features generated from proposal modules such as RPN~\cite{FasterRCNN}.
Since VLMs are pretrained to align images and texts by increasing the similarity of positive image-text pairs and decreasing that of negative pairs.
By combining the detector score with the VLM score, these methods~\cite{F-VLM,CLIPSelf,OV-DQUO,CORA} exploit the zero-shot classification capability of pretrained VLMs for novel-object recognition.
\begin{figure}[t!] 
    \centering
    \includegraphics[width=\linewidth]{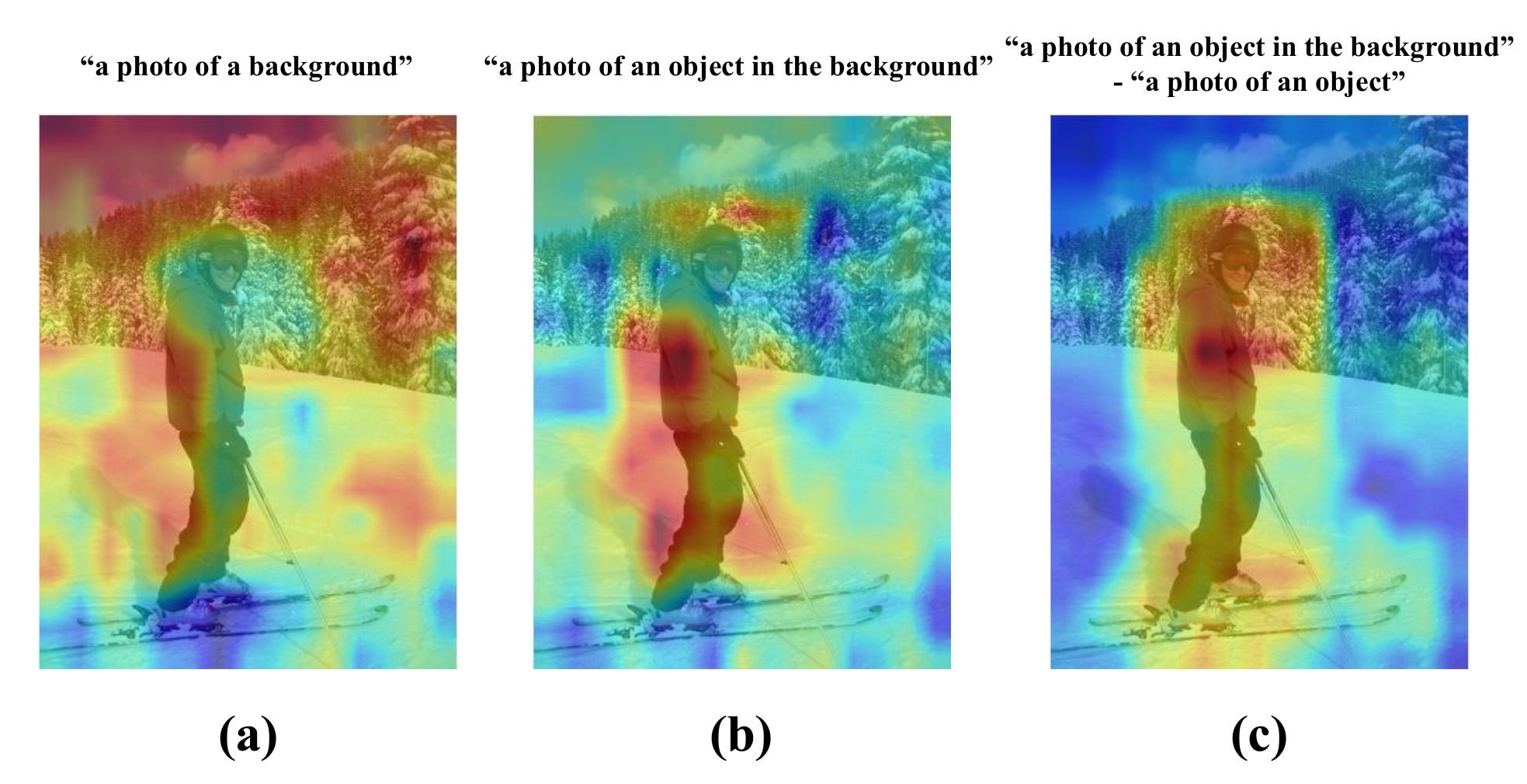} 
    \vspace{-18pt}
    \caption{
    \textbf{Qualitative results of similarities computed by three different prompt strategies.}
    Each visualized heatmap highlights the similarities between cropped images and class-agnostic prompts.
    (a): leveraging the background prompt highlights the background area.
    (b): prompt including both object and background highlights boundary regions of the object.
    (c): leveraging the difference between background and object prompts yields object-containing regions.
}
    
    \label{fig:fig1}
    \vspace{-16pt}
\end{figure}
However, existing F-VLM-style methods still fail to address novel-object detection because the detector is trained with supervision from base classes only.
This training protocol biases the detector toward base-category objects, causing regions containing novel objects to be assigned as background or to receive consistently low detection scores.
As a result, novel-object proposals are often suppressed, which indicates the need for explicit calibration for novel classes.
This issue is related to a broader recognition challenge in which models must remain reliable under distribution shifts, open-set uncertainty, or ambiguous candidate ranking~\cite{ref4, ref5, ref6}.

To compensate for this limitation, we leverage a pretrained VLM kept frozen during detector training, since it preserves generic visual-semantic priors that are not biased by base-class supervision.
Fig.~\ref{fig:fig1} shows 
that pretrained VLMs enable to identify background and foreground regions in an image. By leveraging generic class-agnostic prompts, we verify that pretrained VLMs have following properties.
First, pretrained VLMs recognize background regions. 
As illustrated in Fig.~\ref{fig:fig1}(a), the prompt \textit{``a photo of a background”} produces high similarity for background regions, indicating that pretrained VLMs encode useful cues for background localization and discrimination.
Second, pretrained VLMs recognize regions corresponding to the target concept with reduced reliance on spurious correlations. The contrast between prompts with and without background information tends to highlight the corresponding object regions. As illustrated in Fig.~\ref{fig:fig1}(b), the similarity contrast between the prompt \textit{``a photo of an object in the background”} and \textit{``a photo of an object”} highlights object-containing regions more clearly than either prompt alone, suggesting that generic class-agnostic prompts provide proposal-level localization signals even without category supervision. 
PerceptionCLIP~\cite{PerceptionCLIP} found that using the prompts including contextual attributes reduces reliance on spurious features. Based on this finding,     we hypothesize that the effect in Fig~\ref{fig:fig1}(b)     arises from the contrast between similarities that are differentially influenced by common spurious features from the prompt \textit{``a photo of an object in the background”} and similarities that are more influenced by common spurious features from the prompt \textit{``a photo of an object”}. 
These observations imply that a frozen VLM contains complementary information that is absent from detector scores trained only on base categories. These observations suggest that a proposal can be characterized in a class-agnostic manner by its foreground-aware and background-aware properties. Based on this insight, we propose to recalibrate only the novel category scores using a proposal prior derived from the foreground-aware and background-aware prompt similarities. Rather than modifying detector training or proposal generation, our method acts as an inference-time ranking correction module that suppresses false novel activation on background proposals while promoting better-localized novel proposals in the final scoring stage.
Motivated by these observations, we propose ProCal, a plug-in score recalibration strategy that reduces background-foreground confusion and improves the localization quality of detection scores in existing OVD models without any additional training.

Given the proposals, we extract proposal-level visual features using a frozen pretrained VLM image encoder and compute prompt-based similarities to derive localization-aware and background-aware cues. These cues are integrated into the novel confidence scores. Unlike approaches that require retraining the detector and introducing additional supervision, our method preserves the original OVD architecture and attaches to existing pipelines in a plug-in manner. Our method refines the scoring of generated proposals from RPN so that the confidence becomes better aligned with localization quality.
By calibrating pretrained VLM scores at inference time, our approach improves detection performance for novel classes while remaining training-free.

\noindent Our contributions are summarized as three-fold:
\begin{itemize}
\item We observe that generic class-agnostic prompts in pretrained VLMs provide foreground-background discriminative cues.
\item We propose a training-free plug-in recalibration module that builds a proposal prior from localization-aware foreground and background-aware suppression scores.
\item We show that ProCal improves proposal-level localization-aware ranking while suppressing the confidence of partial novel proposals.
\end{itemize}

\section{RELATED WORKS}

\subsection{Open-Vocabulary Object Detection} 
OVD has advanced primarily by transferring semantic knowledge from vision-language models to the downstream task of object detection. ViLD~\cite{ViLD} and RegionCLIP~\cite{RegionCLIP} improve novel-category recognition by distilling pretrained vision-language knowledge or by learning region-level representations. F-VLM~\cite{F-VLM} shows that a frozen VLM serves as an effective region classifier by combining the detector and VLM outputs at inference time, while CLIPSelf~\cite{CLIPSelf} alleviates the gap between the whole image and the region features through self-distillation. CORA~\cite{CORA} addresses the distribution gap between the whole image and region features and localization difficulty for novel proposals by introducing region prompting and anchor pre-matching. These methods mainly focus on improving region-language alignment, representation quality, or localization generalization. OV-DQUO~\cite{OV-DQUO} 
observes that detectors trained on base categories tend to assign higher confidence to seen classes and to confuse novel objects with background proposals, and mitigates this issue with denoising text-query training and unknown-object supervision by using an open-world detector. 
We introduce a prompt-derived, localization-aware proposal-level prior that reranks existing proposals by favoring object-containing regions while suppressing background regions without modifying detector training or introducing additional unknown supervision. 

\subsection{Vision-Language Models} 
Recent years have witnessed advances in VLMs through contrastive learning. Pioneering works such as CLIP~\cite{clip} and ALIGN~\cite{ALIGN} establish unified image-text embedding spaces by learning cross-modal alignment on web-scale datasets. These models demonstrate remarkable zero-shot transfer capabilities. For instance, CLIP achieves open-vocabulary classification on ImageNet~\cite{ImageNet} with zero-shot setting. This paradigm inspires numerous innovations for downstream tasks including semantic segmentation and object detection. We present an OVD framework that efficiently and effectively leverages pretrained knowledge of VLMs.    Using prompts that encode foreground and background cues at inference time improves novel-object detection performance.

\begin{figure*}[ht!]\vspace{-20pt}
    \centering
        \includegraphics[width=0.8\textwidth]{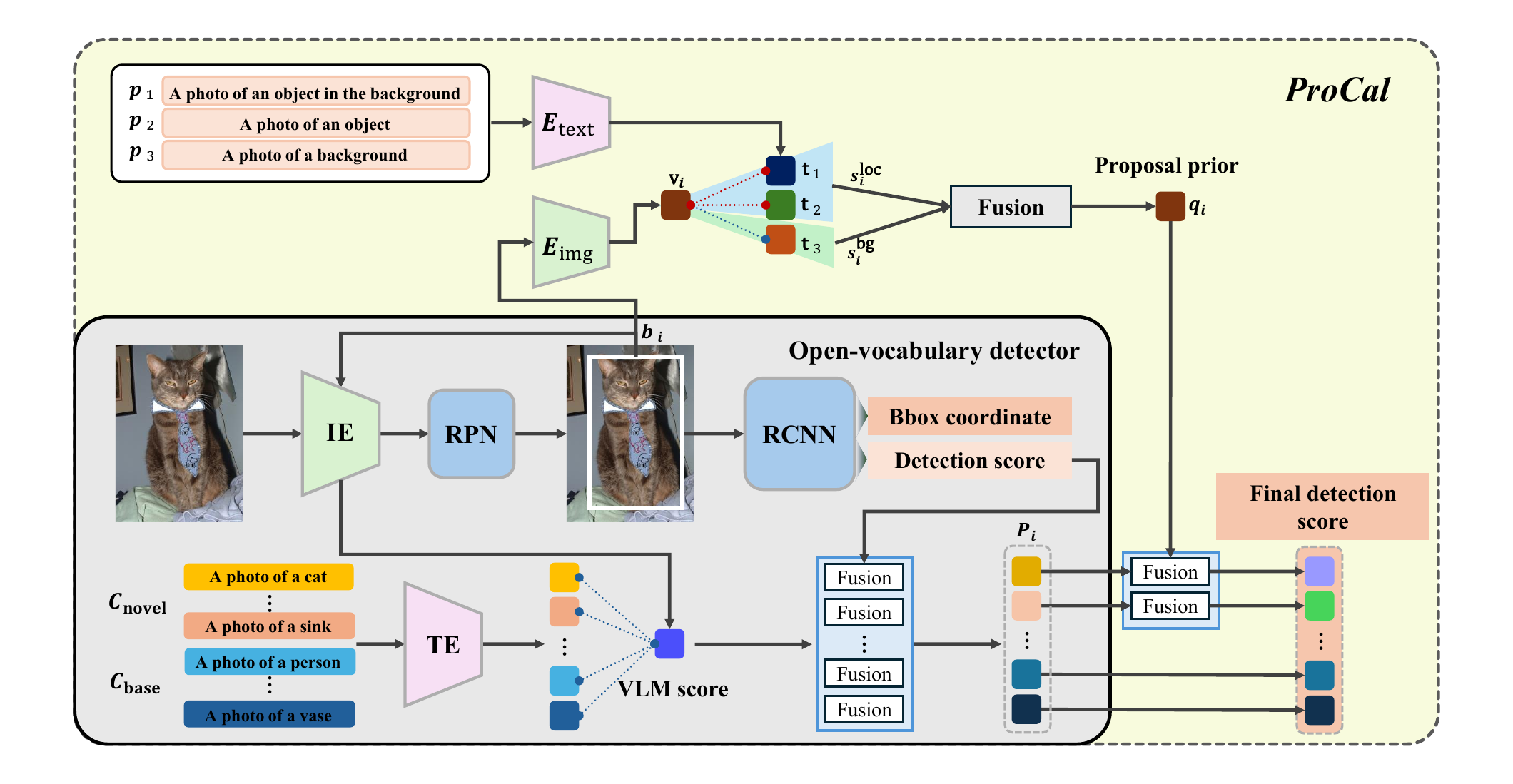} 
        \label{fig1:a}
    \vspace{-13pt}
    \caption{ \textbf{Overview of ProCal}.
    First, ProCal uses region proposals to extract visual features from the VLM.
    After that, ProCal computes localization-aware and background-aware scores for each region.
    Finally, ProCal combines the proposal prior with the proposal-level scores to produce the final detection score. Notably, ProCal does not require any training. 
    }

    \label{fig:figure1}
    \vspace{-14pt}
\end{figure*}

\section{METHOD}


\subsection{Preliminaries}
Given an input image $\mathbf{I} \in \mathbb{R}^{3 \times H \times W}$, a two-stage open-vocabulary detector predicts a set of bounding boxes $\{\mathbf{b}_i\}_{i=1}^{N}$, where $\mathbf{b}_i \in \mathbb{R}^{4}$ together with category scores. ProCal extracts the bounding box $\mathbf{b}_i$ to proposal features. By using the proposal features, ProCal computes localization-aware and background-aware scores. The detector predicts category scores over the test vocabulary $\mathcal{C}_{\text{test}} = \mathcal{C}_{\text{base}} \cup \mathcal{C}_{\text{novel}}$. During training, only the base categories $\mathcal{C}_{\text{base}}$ are annotated, while novel categories $\mathcal{C}_{\text{novel}}$ are introduced only at test time.

\subsection{Proposal Calibration (ProCal)}
Our method is motivated by a key limitation of two-stage OVD: since unseen objects are not explicitly labeled as foreground during detector training, proposals containing novel objects are often confused with true background and receive insufficient confidence at inference. To alleviate this issue, we introduce a proposal-level score recalibration strategy that injects foreground-aware and background-aware cues derived from a frozen pretrained VLM.

Specifically, given the top-$k$ proposals produced by the detector, we extract proposal-level visual features using the frozen VLM image encoder and compute prompt-based similarities with three hand-crafted prompts. From these similarities, we construct (\textit{i}) a localization-aware foreground score and (\textit{ii}) a background-aware suppression score.
These two signals are combined into a proposal prior, which is then fused with the original detection scores for novel categories.
In this way, our method improves the ranking and preservation of novel foreground proposals without modifying the detector architecture, proposal generator, or box regressor.

\subsubsection{Proposal-level features}
Let $\mathcal{B}=\{\mathbf{b}_i\}_{i=1}^{k}$ denote the top-$k$ proposals predicted by the detector. For each proposal $\mathbf{b}_i$, we crop the corresponding image region $\mathbf{I}_{\mathbf{b}_i}$ from the input image and feed it into the VLM image encoder $E_{\mathrm{img}}(\cdot)$.
We extract the proposal feature as follows: 
\begin{equation}
\mathbf{v}_i
=
\frac{E_{\mathrm{img}}(\mathbf{I}_{\mathbf{b}_i})}
{\left\|E_{\mathrm{img}}(\mathbf{I}_{\mathbf{b}_i})\right\|_2}.
\end{equation}
We encode the following three prompts: \textit{``a photo of an object in the background"} as $p_1$, \textit{``a photo of an object"} as $p_2$, and \textit{``a photo of a background"} as $p_3$. We extract the feature of $p_k$ as follows:

\begin{equation}
\mathbf{t}_k
=
\frac{E_{\mathrm{text}}(p_k)}
{\left\|E_{\mathrm{text}}(p_k)\right\|_2},
\qquad k \in \{1,2,3\}\text{,}
\end{equation}
where $E_{\mathrm{text}}(\cdot)$ denotes the frozen text encoder. 

\subsubsection{Localization-aware foreground score}
We aim to capture whether a proposal contains a foreground region rather than a background. Based on the observation that the contrast between $p_1$ and $p_2$ emphasizes object-containing regions, we define the localization-aware score for proposal $\mathbf{b}_i$ as:
\begin{equation}
s_i^{\mathrm{loc}}
=
\phi(\mathbf{v}_i, \mathbf{t}_1)
-
\phi(\mathbf{v}_i, \mathbf{t}_2)\text{,}
\end{equation}
where $\phi(\cdot)$ denotes the cosine similarity. A larger $s_i^{\mathrm{loc}}$ indicates that the proposal contains a foreground object region with meaningful spatial extent.

\subsubsection{Background-aware suppression score}
We measure the extent to which a proposal resembles the background. Since the prompt $p_3$ is responsive to background regions, we define the background score as:
\begin{equation}
s_i^{\mathrm{bg}}
=
\phi(\mathbf{v}_i, \mathbf{t}_3)\text{,}
\end{equation}
where $s_i^{\mathrm{bg}}$ suggests that the proposal corresponds to background or to a low-quality proposal containing background.

\subsubsection{Proposal prior}
To combine the two cues, we first map them into $[0,1]$ using the sigmoid function:
\begin{equation}
\hat{s}_i^{\mathrm{loc}} = \sigma(s_i^{\mathrm{loc}}/ \tau_1),
\qquad
\hat{s}_i^{\mathrm{bg}} = \sigma(-s_i^{\mathrm{bg}}/\tau_2),
\end{equation}
where $\sigma(\cdot)$ denotes the sigmoid function and both $\tau_1$ and $\tau_2$ denote temperature. $\hat{s}_i^{\mathrm{loc}}$ promotes foreground-aware proposals, whereas $\hat{s}_i^{\mathrm{bg}}$ suppresses background proposals. 

We then construct a proposal prior $q_i$ by combining the two terms through a weighted arithmetic mean:
\begin{equation}
q_i
=
{\lambda}\hat{s}_i^{\mathrm{loc}} + 
\left({1-\lambda}\right)\hat{s}_i^{\mathrm{bg}},
\end{equation}
where $\lambda \in [0,1]$ controls the relative contribution of the localization-aware and background-aware cues. 
The design of the proposal prior follows a general cue-fusion principle: weak but complementary signals can yield a more reliable decision variable when aggregated appropriately, a strategy that has been effective in various tasks~\cite{ref7, ref8, ref9}.
These proposal priors reflect whether a proposal contains a foreground object while remaining distinct from the background.
We show ablations of different score combination designs in the experiments section.

\subsubsection{Score fusion}
Let $P_i^j$ denote the original detection score of the baseline detector for proposal $\mathbf{b}_i$ and category $j \in \mathcal{C}_{\text{test}}$. In F-VLM-style OVD frameworks, $P_i^j$ is obtained by combining detector confidence and VLM-based category similarity. We recalibrate the original detection score using the  $q_i$.
Since our goal is to mitigate the underestimation of novel objects, we apply the recalibration only to novel categories while leaving base-category scores unchanged:
\begin{equation}
\tilde{P}_i^j
=
\begin{cases}
P_i^j, & j \in \mathcal{C}_{\text{base}}, \\[6pt]
\left(P_i^j\right)^{1-\gamma}
\left(q_i\right)^{\gamma}, & j \in \mathcal{C}_{\text{novel}},
\end{cases}
\label{eq8}
\end{equation}
where $\gamma\in[0,1]$ controls the strength of the proposal-level score recalibration.
This fusion preserves the original category-specific evidence from the detector while injecting a proposal-level prior that reflects foreground-background separability. Consequently, novel proposals that contain foreground objects are promoted, whereas background-aware or low-quality proposals are relatively suppressed.

\subsection{Training and Inference}

\subsubsection{Training}
Our method introduces no additional training objective and preserves the training procedure of the detector. The detector is trained in the standard open-vocabulary setting using only annotations from $\mathcal{C}_{\text{base}}$. The pretrained VLM image encoder and text encoder used in our method remain frozen throughout both training and inference.

\subsubsection{Inference}
At inference time, the baseline detector first generates proposals and predicts the original detection scores $P_i^j$. We then select the top-$k$ proposals, crop each proposal region, and extract proposal-level visual features using the frozen VLM image encoder.
We use a frozen VLM different from that of the baseline to extract proposal-level visual features.  
Prompt-based similarities are computed to obtain the localization-aware score $s_i^{\mathrm{loc}}$ and the background-aware score $s_i^{\mathrm{bg}}$, which are further combined into the  $q_i$. Finally, we recalibrate the detection scores of novel categories using Eq.~\eqref{eq8}, while keeping base-category scores unchanged.

Notably, the proposed method does not alter proposal generation or box regression. Instead, it acts as an inference-time calibration module that improves the ranking and preservation of surviving novel proposals under foreground-background ambiguity.

\section{EXPERIMENTS}

\begin{table}[!t]

	\centering
	\scriptsize
	\caption{\textbf{Performance comparison with state-of-the-art methods on the OV-COCO dataset.}}
    \vspace{-5pt}
    \label{tab:COCO}
    \resizebox{\linewidth}{!}{
	\begin{tabular}{l|c|ccc}
        \toprule
        \scalebox{1.0}{\textbf{Method}}    & \scalebox{1.0}{\textbf{Backbone}} &\scalebox{1.0}{$\text{AP}_{\text{Base}}^{50}$}     &\scalebox{1.0}{$\text{AP}_{\text{Novel}}^{50}$}     &\scalebox{1.0}{$\text{AP}^{50}$}  \\
        
        \midrule
        
        {RegionCLIP} & \multirow{4}*{RN50} &57.1  &31.4  &50.4  \\
        Detic   & & 47.1   & 27.8  & 45.0  \\

        F-VLM &  &-  &28.0  &39.6    \\

        {CORA}  & &35.5 &\underline{35.1} &35.4  \\
        \midrule
        
        {RegionCLIP} &\multirow{3}*{RN50x4} &61.6  &39.3  &55.7    \\
        {CORA}  & & 44.5 &41.7 &43.8  \\
        CORA+  & &60.9 &\underline{43.1} &56.2  \\
        \midrule
        {DST-Det} & \multirow{5}*{ViT-B/16} &59.6  &\underline{41.3}  &54.8  \\
        {OV-DQUO} & & 42.6  &39.4   & 41.7  \\
        {CLIPSelf + F-ViT} &  & 54.8  &37.4  &50.3  \\
        {CLIPSelf + NoOVD} &  &56.5  &37.9  &51.6   \\
        {CLIPSelf + F-ViT + \textbf{ProCal}} & 
         & 54.3  &\textbf{40.1 (+2.7)}  &50.5  \\
        \midrule
        {DST-Det} & \multirow{5}*{ViT-L/14} &  61.9  &\underline{46.7}  &58.0   \\
        OV-DQUO &  & 48.3  &45.3  &47.5  \\ 
        {CLIPSelf + F-ViT} &  &64.2  &44.5  &59.0    \\
        
        {CLIPSelf + NoOVD} &  &64.7  &45.4  &59.7   \\
	      {CLIPSelf + F-ViT + \textbf{ProCal}} &  &63.4  &\textbf{45.7 (+1.2)} &58.8    \\
        
	\bottomrule    
	\end{tabular}
    }
    
\vspace{-15pt}
\end{table}

\subsection{Datasets and Evaluation Metrics}
We evaluate our method on the COCO~\cite{COCO} and LVIS~\cite{LVIS}. By following OVR-CNN~\cite{OVRCNN}, we divide the 65 classes in the COCO dataset into 48 base classes and 17 novel classes. In the COCO benchmark, we report the box mean average precision at an IoU threshold of 0.5 for 65 classes, base classes, and novel classes, denoted as $\text{AP}^\text{50}$, $\text{AP}_{\text{base}}^\text{50}$ and $\text{AP}_{\text{novel}}^\text{50}$, respectively.
ProCal is evaluated on LVIS comprising 100K images and 1,203 categories. The categories are split into three groups: common, frequent, and rare classes. We report the box mean average precision from IoU threshold 0.5 to 0.95 for rare classes and all classes, denoted as $\text{AP}_r$ and $\text{AP}$, respectively.

\subsection{Comparison Methods}
We select OVD methods employing transfer learning such as F-VLM~\cite{F-VLM}, CLIPSelf~\cite{CLIPSelf}, and OV-DQUO~\cite{OV-DQUO}, knowledge distillation such as RegionCLIP~\cite{RegionCLIP}, and pseudo-labeling such as Detic~\cite{Detic} and CORA~\cite{CORA} for comparison. 
NoOVD~\cite{No-OVD} improves discovery for novel objects via knowledge distillation and applies to the top of CLIPSelf. We set our baseline as CLIPSelf ViT-B/16 and ViT-L/14 on the open-vocabulary benchmarks. 

\subsection{Implementation Details}
We use OpenCLIP~\cite{OpenCLIP} ViT-B/16 model as image and text encoder to extract proposal-level features for the proposal prior. For the proposal prior fusion, we use $\lambda=0.5$, $\tau_1=0.02$, and $\tau_2=0.1$. For final score fusion, we set $\gamma=0.2$.
We set the non-maximum suppression (NMS) threshold to 0.3 and 0.5 on OV-COCO and OV-LVIS, respectively.
Other settings are the same as the CLIPSelf baseline.

\subsection{Quantitative Results}

\subsubsection{Results on OV-COCO}
Tab.~\ref{tab:COCO} presents the results of other OVD methods and ProCal on the OV-COCO. 
Although ProCal requires no additional training or extra data, it performs comparably to or better than other OVD methods that relied on additional training process or additional datasets.

\subsubsection{Results on OV-LVIS}
Tab.~\ref{tab:LVIS} presents the results of other OVD methods and ProCal on OV-LVIS. 
ProCal with ViT-L/14 performs comparable $\text{AP}_r$ while higher $\text{AP}$ to OV-DQUO. These results support that our simple localization-aware foreground and background suppression method operates properly for novel object detection.


\begin{table}[t!]
	\centering
	\caption{\textbf{Performance comparison with state-of-the-art methods on the OV-LVIS dataset.}} 
    \label{tab:LVIS}
    \vspace{-5pt}
    \resizebox{0.84\linewidth}{!}{
	    \begin{tabular}{l|c|cc}
        \toprule
        \textbf{Method}    & \textbf{Backbone}  &$\text{AP}_\text{r}$ &$\text{AP}$  \\\midrule
        {RegionCLIP} & \multirow{3}*{RN50} & 17.1 & 28.2\\
        Detic  &  &\underline{24.9} &32.4 \\
        {F-VLM} & & 18.6 &24.2\\
        \midrule
        {RegionCLIP} & \multirow{4}*{RN50x4} &  22.0 & 32.3\\
        CORA &  &22.2  & -\\
        CORA+ &  &\underline{28.1} & - \\
        {F-VLM} & & 26.3  &28.5\\
        \midrule
        DST-Det &   & 26.2  &- \\
        {OV-DQUO} &   &\underline{29.4}  &26.5\\
        {CLIPSelf + F-ViT} &   &25.8 &26.4\\
        {CLIPSelf + NoOVD}  &   & 28.3  &27.6\\
        {CLIPSelf + F-ViT + \textbf{ProCal}} &   &\textbf{26.9 (+1.1)} & 26.0  \\
        \midrule
        {OWL-ViT} & \multirow{5}*{ViT-L/14}  & 31.2 &34.6\\
        {OV-DQUO} &  &\underline{39.5} &33.7 \\ 
	     {CLIPSelf + F-ViT}  &   &36.8  &37.3 \\
        {CLIPSelf + NoOVD}&   & 37.2 & 37.7\\      
	    {CLIPSelf + F-ViT + \textbf{ProCal}}  &   &\textbf{39.3 (+2.5)}  &37.3 \\
        \bottomrule    
    	\end{tabular}
        }
\vspace{-15pt}
\end{table}

\subsection{Analysis of Localization-Aware Score Refinement}
\subsubsection{Localization-aware improvement across IoU thresholds and object scales}
Tab.~\ref{tab:localization-aware} analyzes the effect of improving the localization quality of the detection score. In Tab.~\ref{tab:localization-aware}(a), the improved performance is not restricted to $\text{AP}^{50}_{\text{Novel}}$. We observe consistent improvements at higher IoU thresholds, including $\text{AP}_{\text{Novel}}^{75}$ and $\text{AP}^{90}_{\text{Novel}}$. The results suggest that our localization-aware foreground score elevates well-localized novel proposals rather than merely increasing the overall confidence of novel proposals.
In Tab.~\ref{tab:localization-aware}(b), ProCal is particularly effective for medium-sized novel objects, where the region features retain sufficient object semantics while still suffering from substantial misalignment between confidence and localization quality. This result suggests that ProCal mainly improves the ranking of novel proposals.
\subsubsection{Proposal ranking quality analysis}
Tab.~\ref{tab:topk_analysis} analyzes the composition of top-ranked proposals.
To better understand why the proposed refinement improves the novel detection performance, we analyze whether the final detection score better reflects the localization quality of the proposal.
We divide proposals into three groups: 
(1) novel positive proposals: IoU $\geq 0.5$ with novel ground-truth boxes, (2) partial novel proposals: $0.1 \leq$ IoU $<0.5$ with novel ground-truth boxes, and (3) background proposals: IoU $< 0.1$ with base and novel ground-truth boxes.
We then compare the original and refined detection scores with respect to proposal-level ranking quality, defined as whether true novel proposals are ranked above background and partial proposals.

After refinement, the proportion of novel positive proposals increases consistently across all top-$k$ sets, while the proportion of background proposals decreases.
These results show that the upper-ranked novel predictions become cleaner after refinement.
This trend is consistent with candidate-selection and localization-refinement studies~\cite{ref6, ref10}, where reliable intermediate candidates are important for final prediction quality. In our setting, we observe that proposal-level score calibration improves localization-aware novel-object ranking.  
Therefore, the proposed method improves the localization-aware ranking quality of detections, which explains the observed improvement in novel AP.
\begin{table}[!t]
    \caption{\textbf{Quantitative comparison of different AP settings against the baseline.} We use the OV-COCO dataset and ViT-B/16.}
    \vspace{-5pt}
    
    \begin{subtable}[h]{0.24\textwidth}
    \centering
    \caption{}
    \vspace{-5pt}
    \resizebox{0.98\linewidth}{!}
    {
        \renewcommand{\arraystretch}{1.3}
        \begin{tabular}{c|cc}
        \toprule
        \ & \makecell{CLIPSelf \\ + F-ViT}
        & \makecell{CLIPSelf \\ + F-ViT + ProCal} \\
        \midrule
        {$\text{AP}_{\text{Novel}}^{50:95}$} & \scalebox{1.0}{17.9} & \scalebox{1.0}{\textbf{19.2 (+1.3)}} \\
        {$\text{AP}^{50}_{\text{Novel}}$} & \scalebox{1.0}{37.4} & \scalebox{1.0}{\textbf{39.9 (+2.5)}}\\
        {$\text{AP}^{75}_{\text{Novel}}$} & \scalebox{1.0}{15.3} & \scalebox{1.0}{\textbf{16.8 (+1.5)}} \\
        {$\text{AP}^{90}_{\text{Novel}}$} & \scalebox{1.0}{1.2} & \scalebox{1.0}{\textbf{1.4 (+0.2)}} \\
        \bottomrule
        \end{tabular}
    }
    \label{tab:localization-aware(a)}
    \end{subtable}
    \hfill
    \begin{subtable}[h]{0.24\textwidth}
    \centering
    \caption{}
    \vspace{-5pt}
    \resizebox{0.98\linewidth}{!}
    {
        \renewcommand{\arraystretch}{1.3}
        \begin{tabular}{c|cc}
        \toprule
        \ & \makecell{CLIPSelf \\ + F-ViT}
        & \makecell{CLIPSelf \\ + F-ViT + ProCal} \\
        \midrule
        {$\text{AP}^{\text{Small}}_{\text{Novel}}$} & \scalebox{1.0}{13.0} & \scalebox{1.0}{\textbf{13.9 (+0.9)}}\\
        {$\text{AP}^{\text{Medium}}_{\text{Novel}}$} & \scalebox{1.0}{21.2} & \scalebox{1.0}{\textbf{23.8 (+2.6)}} \\
        {$\text{AP}^{\text{Large}}_{\text{Novel}}$} & \scalebox{1.0}{24.6} & \scalebox{1.0}{\textbf{24.8 (+0.2)}} \\
        \bottomrule
        \end{tabular}
    }
    \label{tab:localization-aware(b)}
    \end{subtable}
    \label{tab:localization-aware}
\end{table}

\begin{table}[!t]
\centering
\caption{\textbf{Analysis of the three different top-ranked proposals.}
We include novel positive, partial novel, and background proposals.
We refine the proposals using top-$k$ and report the accuracy measured by varying $k$ using the OV-COCO dataset with ViT-B/16.
}
\label{tab:topk_analysis}
\resizebox{\linewidth}{!}{
\begin{tabular}{c|cc|cc|cc}
\toprule
Top-$k$ 
& \multicolumn{2}{c|}{Novel positive proposals (\%)} 
& \multicolumn{2}{c|}{Partial novel proposals (\%)} 
& \multicolumn{2}{c}{Background proposals (\%)} \\
& \multicolumn{2}{c|}{$\text{IoU}\geq 0.5$} & \multicolumn{2}{c|}{$0.1 \leq \text{IoU}<0.5$} & \multicolumn{2}{c}{$\text{IoU}<0.1$} \\ 
 & Original & Refined & Original & Refined & Original & Refined \\
\midrule
10  & 24.7 & \textbf{25.8} & 5.7 & \textbf{5.4} & 9.5 & \textbf{8.0} \\
20  & 22.5 & \textbf{23.5} & 6.3 & \textbf{6.2} & 10.6 & \textbf{9.1} \\
50  & 17.5 & \textbf{18.2} & 7.5 & 7.6 & 13.6 & \textbf{12.2} \\
100 & 12.9 & \textbf{13.3} & 7.8 & 8.0 & 18.5 & \textbf{17.2} \\
\bottomrule
\end{tabular}
}
\end{table}

\begin{table}[!t]
    \centering
    \caption{\textbf{Ablation study of the proposed ProCal on the OV-COCO dataset.} All results are based on ViT-B/16.}
    \vspace{-5pt}
    \resizebox{0.75\linewidth}{!}{
    \begin{tabular}{c|cc|cc}
    \toprule
    {\ \#\ }  & \makecell{\ \ \ \ $s^{loc}\ \ \ \ $}
    & \makecell{\ \ \ \ $s^{bg}$\ \ \ \ }
    & \scalebox{1.0}{$\text{AP}^{50}_{\text{Novel}}$}
    & \scalebox{1.0}{$\text{AP}^{50:95}_{\text{Novel}}$} \\
    \midrule
    1 & \scalebox{1.0}{-} & \scalebox{1.0}{-} & \scalebox{1.0}{37.4} & \scalebox{1.0}{17.9} \\
    2 & \scalebox{1.0}{\ding{51}} & \scalebox{1.0}{\ding{55}} & \scalebox{1.0}{40.0} & \scalebox{1.0}{19.2}\\
    3 & \scalebox{1.0}{\ding{55}} & \scalebox{1.0}{\ding{51}} & \scalebox{1.0}{39.5} & \scalebox{1.0}{19.0} \\
    4 & \scalebox{1.0}{\ding{51}} & \scalebox{1.0}{\ding{51}} & \scalebox{1.0}{\textbf{40.1}} & \scalebox{1.0}{\textbf{19.3}} \\
    \bottomrule  
    \end{tabular}
    }
    \label{tab:ablation study}
\vspace{-8pt}
\end{table}

\begin{table}[!t]
    \caption{\textbf{Ablation study of score fusion for the proposal prior.} We study different score fusion mechanisms of ProCal. We report {$\text{AP}^{50}_{\text{Novel}}$} on the OV-COCO with ViT-L/14.}
    \vspace{-5pt}
    \begin{subtable}[h]{0.49\linewidth}
    \centering
    \caption{Geometric fusion}
    \vspace{-5pt}
    \resizebox{0.9\linewidth}{!}{
    \begin{tabular}{c|cc|c}
    \toprule
     \makecell{$\lambda$} & \makecell{$\tau_1$} & \makecell{$\tau_2$}
    & \makecell{$\text{AP}^{50}_{\text{Novel}}$} \\
    \midrule
     0.5 & 0.04 & 0.1 & 45.4 \\
    0.5 & 0.04 & 0.04 & 45.4 \\
      {0.5} & {0.02} & {0.1} & {45.6} \\
      0.5 & 0.02 & 0.04 & 45.4 \\
    
    \bottomrule
    \end{tabular}
    }
    \end{subtable}
    \hfill
    \begin{subtable}[h]{0.49\linewidth}
    \centering
    \caption{Arithmetic fusion}
    \vspace{-5pt}
    \resizebox{0.9\linewidth}{!}{
    \begin{tabular}{c|cc|c}
    \toprule
    \makecell{$\lambda$} & \makecell{$\tau_1$} & \makecell{$\tau_2$}
    & \makecell{$\text{AP}^{50}_{\text{Novel}}$} \\
    
    \midrule
    0.5 & 0.04 & 0.1 & 45.4 \\
     0.5 & 0.04 & 0.04 & 45.2 \\
     0.5 & 0.02 & 0.1 & 45.7 \\
      0.5 & 0.02 & 0.04 & 45.1 \\
    \bottomrule
    \end{tabular}
    }
    \end{subtable}
    \label{tab:ablation study of score fusion}
    \vspace{-21pt}
\end{table}

\subsection{Ablation Study}
\subsubsection{Ablation study on main components}
We perform an ablation study to analyze the effects of the localization-aware foreground score and the background-aware suppression score. As reported in Tab.~\ref{tab:ablation study}, the foreground score contributes more substantially to the overall performance improvement. This indicates that object-aware localization cues are critical for boosting novel AP. Meanwhile, the suppression score helps mitigate the tendency of background proposals to be incorrectly classified as novel.
\subsubsection{Score fusion for proposal prior}

Tab.~\ref{tab:ablation study of score fusion} analyzes the effect of score fusion mechanisms and hyperparameter choices in ProCal. Although arithmetic fusion achieves the best performance, ProCal remains consistently effective across both arithmetic and geometric methods as well as different $\tau_1$ and $\tau_2$ settings. The performance variation is small across all configurations, showing that the proposed method is robust to the choice of fusion strategy and hyperparameters. Based on the best result, we use arithmetic fusion as the default setting in ProCal method.

\subsection{Qualitative Results}
\begin{figure}[!t]
    \centering
    \includegraphics[width=\linewidth]{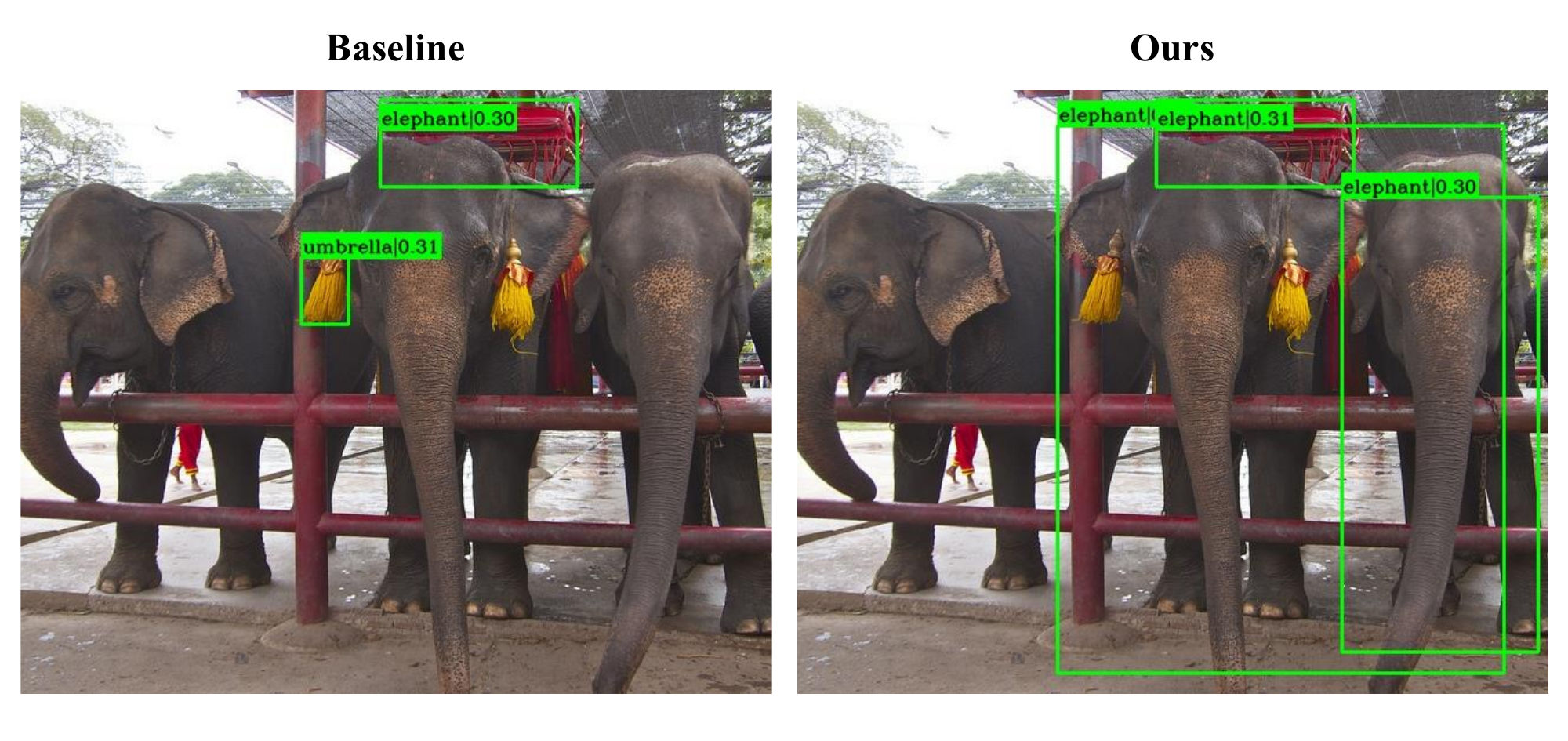} 
    \vspace{-18pt}
    \caption{
    \textbf{Visualized detection results on the OV-COCO dataset with ViT-B/16.}
    We highlight the detected novel classes using a green box.
    }
    \label{fig:vis_bbox}
    \vspace{-12pt}
\end{figure}

We present detection results for the baseline and ProCal after NMS with a threshold of 0.3 on the OV-COCO validation sets. 
In Fig.~\ref{fig:vis_bbox}, the baseline predicts a partial object box for the elephant region and misclassifies a part of the elephant region as an umbrella.
Compared with the baseline, ProCal assigns higher confidence scores to boxes that more heavily overlap the ground-truth boxes of elephants in the image.
We observe that ProCal suppresses false novel proposals as well as background proposals.

\section{CONCLUSION}

In this paper, we addressed a key limitation of two-stage open-vocabulary object detection. Unseen objects are not explicitly learned as valid foreground during training. Surviving novel proposals are often ranked below background proposals or low-quality proposals at inference. To mitigate this issue, we proposed a simple inference-time score recalibration method that injects foreground-aware and background-aware cues from a frozen pretrained VLM into the final scoring process. 
Our analyses showed that ProCal improves proposal-level ranking quality. ProCal suppresses false novel activation and more consistently ranks true novel proposals above background proposals and partial novel proposals. These results suggest that localization-aware score calibration is an effective direction for improving novel object detection, and that frozen VLMs provide useful class-agnostic cues for correcting proposal-level ranking bias.

\bibliographystyle{IEEEtran}
\bibliography{REFERENCE}
\end{document}